\pdfoutput=1

\documentclass[11pt]{article}

\usepackage[]{acl}

\usepackage{times}
\usepackage{latexsym}
\usepackage{booktabs}
\usepackage{multirow}
\usepackage{amsmath}
\usepackage{pifont}
\usepackage{hyperref}

\usepackage[T1]{fontenc}

\usepackage[utf8]{inputenc}

\usepackage{microtype}

\usepackage{graphicx} 
\usepackage{color}

\usepackage{amssymb}
\usepackage{pifont}

\newcommand{\cmark}{\ding{51}}%
\newcommand{\xmark}{\ding{55}}%

\definecolor{user_persona}{RGB}{255,126,121}
\definecolor{bot_persona}{RGB}{0,150,255}
\title{Long Time No See! Open-Domain Conversation with Long-Term Persona Memory}

\newcommand\Mark[1]{\textsuperscript{#1}}

\newcommand*\samethanks[1][\value{footnote}]{\footnotemark[#1]}

\author{Xinchao Xu\Mark{1}\thanks{\quad Equal contribution. The work was done when Zhibin Gou and Shihang Wang were doing internship at Baidu.} , Zhibin Gou\Mark{1,2}\samethanks\ , Wenquan Wu\Mark{1}, Zheng-Yu Niu\Mark{1}, \\
        {\bf Hua Wu\Mark{1}, Haifeng Wang\Mark{1} \and Shihang Wang\Mark{3}} \\
        \Mark{1}Baidu Inc., China \\
        \Mark{2}School of Computer Science, Beijing University of Posts and Telecommunications \\
        \Mark{3}Columbia University \\
        \texttt{\{xinchaoxu,wuwenquan01,niuzhengyu,wu\_hua,wanghaifeng\}@baidu.com}\\
        \texttt{zebgou@gmail.com}, \texttt{sw3275@columbia.edu}}

\begin{document}
\maketitle


\begin{abstract}
Most of the open-domain dialogue models tend to perform poorly in the setting of long-term human-bot conversations. The possible reason is that they lack the capability of understanding and memorizing long-term dialogue history information. To address this issue, we present a novel task of \textbf{L}ong-t\textbf{e}rm \textbf{M}emory C\textbf{on}versation (LeMon) and then build a new dialogue dataset DuLeMon and a dialogue generation framework PLATO-LTM with a Long-Term Memory (LTM) mechanism. This LTM mechanism enables our system to accurately extract and continuously update long-term persona memory without requiring multiple-session dialogue datasets for model training.
To our knowledge, this is the first attempt to conduct real-time dynamic management of persona information of both parties, including the user and the bot. Results on DuLeMon indicate that PLATO-LTM can significantly outperform baselines in terms of long-term dialogue consistency, leading to better dialogue engagingness \footnote{Our data and codes are released at \url{https://github.com/PaddlePaddle/Research/tree/master/NLP/ACL2022-DuLeMon}}.

\end{abstract}

\section{Introduction}


Persona is crucial for open-domain dialogue systems to establish long-term intimacy with users \cite{10.1145/3383123}. Existing persona dialogue datasets such as PersonaChat \citep{zhang-etal-2018-personalizing, DBLP:journals/corr/abs-1902-00098} and models \citep{DBLP:conf/acl/LiGBSGD16, DBLP:journals/corr/ZhangLWZ17, ijcai2018-595} have greatly facilitated the chatbot with configurable and persistent personalities. 



\begin{figure}[t]
  \centering
    \includegraphics[width=8cm]{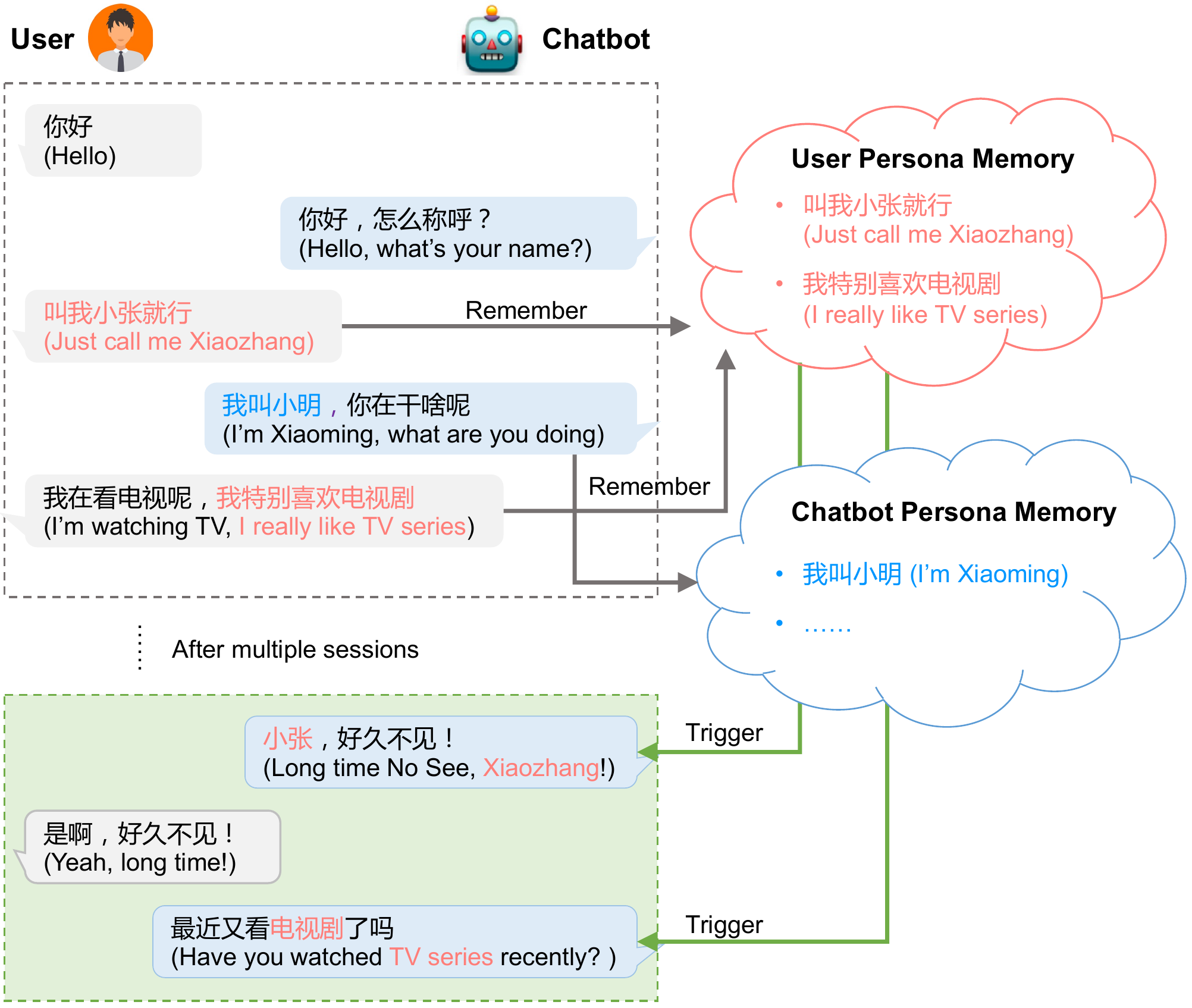}
    \caption{A sample of long-term conversation with memory. 
    At first, the chat partner is not familiar with each other, so the goal is to get to know each other; Then, after multiple sessions, the chatbot already has a certain understanding and memory of the user's persona and its own persona, making the deep chat possible.}
    \label{fig:1}
\end{figure}

Nevertheless, current open-domain dialogue systems still cannot build a long-term connection with humans. 
The possible reason is that they lack the capability of understanding and memorizing long-term dialogue history information, which we called \textbf{long-term persona ability}.
Remembering and actively utilizing the user's persona increases engagingness and contributes to long-term friendships between chatbot and user \cite{10.5555/3237383.3237945}.
Without this ability, the current state-of-the-art models, such as Meena \cite{DBLP:journals/corr/abs-2001-09977},  Blender \cite{DBLP:conf/eacl/RollerDGJWLXOSB21}, and  PLATO \cite{DBLP:journals/corr/abs-2006-16779}, tend to talk to people like strangers in long-term conversations.

Despite the importance and challenge of utilizing long-term persona in open-domain dialogue, as far as we know, the long-term persona ability of large-scale models is less studied due to a lack of both task design and corresponding dataset.
Previous long-term persona dialogue systems \cite{DBLP:conf/interspeech/KimBCRKL14, 7072837} are mainly rule-based systems without large-scale pre-training models, in which researchers proposed various episodic memory architectures to extract, store and manage relevant facts in prior interactions for use in future dialogs \cite{10.5555/3237383.3237945}.

In addition, existing persona conversation datasets \citep{zhang-etal-2018-personalizing, DBLP:journals/corr/abs-1902-00098, DBLP:journals/corr/abs-1901-09672} focus only on the consistency of the chatbot's own persona and ignore the memory and utilization of the user's persona. And they all set fixed persona that cannot be updated during the chat.
Recently, \citet{xu2021goldfish} proposed MSC dataset as a multi-session extension of PersonaChat, and its sessions are additionally annotated with summaries of important personal points. Similar to the previous episodic memory architecture, \citet{xu2021goldfish} summarize and recall previous conversations for future dialogue generation. 
The stored documents in MSC will not be dynamically modified and will increase infinitely as the conversation progresses. Furthermore, the retrieval-augmented generative models rely on a long-session conversation dataset for training, which is expensive and difficult to annotate. 

To address the limitations of existing models and the above issues, we defines the \textbf{LeMon} (\textbf{L}ong-t\textbf{e}rm \textbf{M}emory C\textbf{on}versation) task and propose a new dataset named \textbf{DuLeMon}, which focuses not only on the consistency of the bot's own persona but also on the active construction and utilization of the user's persona in a long-term interaction (ie. mutual persona).
We demonstrate an example dialogue in DuLeMon in Figure \ref{fig:1}. In DuLeMon, we assume that the two speakers have previously interacted with each other and that the chatbot remembers part of the user’s persona. Besides, both the user and chatbot grounding persona are annotated in each utterance.


Based on our collected dataset, we carefully design a novel \textbf{PLATO-LTM} framework for the long-term persona dialogue setting by adding a plug-and-play long-term memory (LTM) to the state-of-the-art open-domain dialogue model \citep{DBLP:journals/corr/abs-2006-16779}. It enables us to study long-term persona conversations without relying on the long-session dataset.
PLATO-LTM can extract both parties' persona information from the conversation in real time, write it to persona memory respectively, and retrieve both parties' persona information from memory to generate responses. The PLATO-LTM framework consists of three modules:
(1) Persona Extractor (PE): The memory is updated by filtering irrelevant information and extracting persona sentences through a classifier. 
(2) Long-Term Memory (LTM): Two separated long-term memories store the explicit persona information of interlocutors. 
(3) Generation Module: We use the large-scale model and the retrieved persona sentences of the user and chatbot are directly concatenated with dialogue context as model input.

Our major contributions are as follows:

\begin{itemize}
\item [(1)] We firstly propose the long-term persona chat task LeMon for Chinese long-term conversations.
Our proposed DuLeMon dataset is also the largest multi-turn Chinese mutual persona chat dataset currently available.
\item [(2)] We proposed a PLATO-LTM framework that extracts and remembers both user’s and the chatbot's persona in real time, enabling the chatbot to have long-term persona dialogue without training on long-session data.
\item [(3)] Automatic and human evaluation show that our method significantly improves the consistency of the state-of-the-art in long conversations, making the response more engaging while ensuring coherency.
\end{itemize}

\begin{table*}
\centering
\small
\begin{tabular}{llllll}
\toprule
\textbf{Dataset} & \textbf{Persona} & \textbf{Mutual}&\textbf{\# Dialogues} & \textbf{Language} & \textbf{Multi-turn} \\
\midrule
PersonaChat \citep{zhang-etal-2018-personalizing} & Text & \xmark & 10,907 & English& Yes \\
PersonalDialog \citep{DBLP:journals/corr/abs-1901-09672} & Structure & \xmark & 20,830,000 & Chinese & part \\
XPersona \citep{DBLP:journals/corr/abs-2003-07568} & Text & \xmark & 16,878 & Multilingual & Yes \\
PEC \citep{zhong-etal-2020-towards} & Text & \xmark &355,000 & English & Yes \\
PCR \citep{mazare-etal-2018-training} & Text & \xmark & 700,000,000 & English & Yes \\
MSC \citep{xu2021goldfish} & Text & \cmark & 5,001 & English & Yes \\
\midrule
DuLeMon (Ours) & Text & \cmark & 27,501 & Chinese & Yes \\
\bottomrule
\end{tabular}
\caption{\label{dataset-table}
Comparison of our dataset DuLeMon with other datasets.
}
\end{table*}

\section{Related Work}
\noindent \textbf{Persona Dialogue:}
As described in \citet{10.1145/3383123}, there is much work related to persona dialogue. Generally speaking, these works can be divided into implicit persona models and explicit persona models. In the implicit model, the persona is represented in the form of the semantic persona vector. \citet{DBLP:conf/interspeech/KimBCRKL14} proposed a retrieval-based method to integrate persona and user interests into the dialogue system. Because these models are implicit methods, they are not easy to interpret and control in target response generation. In \citet{ijcai2018-595}, an explicit persona model is proposed to generate consistent responses for given persona information. The persona information of the machine includes name, gender, hobbies, and so on. In this way, the given persona information can be better used for model generation. There are also many persona chat datasets that have been constructed to develop models, as shown in Table \ref{dataset-table}. 
In particular, the introduction of the PersonaChat \citep{zhang-etal-2018-personalizing, DBLP:journals/corr/abs-1902-00098} dataset has extensively promoted the development of this field where the crowd-workers are simply asked to "chat with the other person naturally and try to get to know each other." However, the user's persona was unknown to the bot, so the dialogue was like strangers exchanging information.
In contrast, our proposed DuLeMon dataset requires the chatbot to actively remember and use the user's persona to improve conversational engagements and increase the intimacy between interlocutors in long-term interactions.

\noindent \textbf{Dialogue Model with External Memory:} As described in \citet{Lim2012}, there are various memory models used by the rule-based dialogue systems.  In \citet{7072837}, user-related information is memorized and used to rewrite the response. In \citet{ElvirGonzalezWallsWilder+2017+1+21}, a unified episodic memory architecture for Embodied Conversational Agents (ECAs) is proposed. They describe a process that determines the prevalent contexts in the conversations obtained from the interactions. In \citet{10.5555/3237383.3237945}, the authors introduce an agent that uses its conversational memory to revisit shared history with users to maintain a coherent social relationship over time. However, they find it challenging to leverage the shared history with individual users and hard to accommodate expected conversational coordination patterns. Apart from studies in rule-based dialogue systems mentioned above,  \citet{xu2021goldfish} shows how large-scale pre-training generative dialogue models trained on existing datasets perform poorly in the long-term conversation setting and proposes a new extended English conversation dataset, entitled Multi-Session Chat (MSC). Different from them, our novel dataset DuLeMon does not rely on long sessions with high collection costs to study long-term memory problems in the persona chat, with significant differences in task design and data collection.

\begin{figure*}[htp]
    \centering
    \includegraphics[width=16cm]{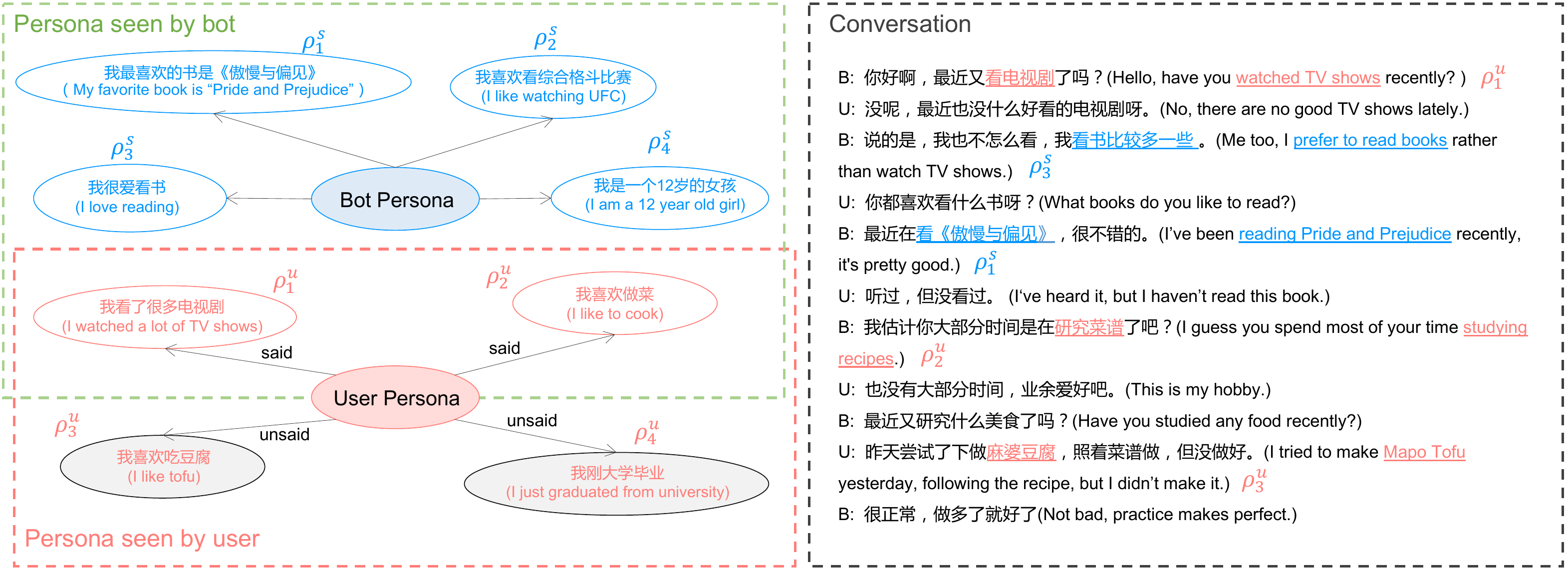}
    \caption{Example of our proposed DuLeMon dataset with both chatbot's and user’s persona. It has two important features: one is that during the conversation, the chatbot can see the persona of both parties; the other is that the persona information associated with the response is explicitly labeled in our dataset which is shown as the $\rho^u$ and $\rho^s$ in the figure.}
    \label{fig:3}
\end{figure*}

\section{Data Collection}

\textbf{Task Definition}. Given dialogue context $c =\{u_1, s_1, u_2, s_2, ..., u_{t-1}, s_{t-1}, u_{t}\}$ , where $u$ and $s$ represent the user and the chatbot respectively. Each speaker has its corresponding persona description that consists of a set of sentences, we define the user persona as $\rho^u = \{\rho^u_1, \rho^u_2, . . . , \rho^u_m\}$, and the chatbot persona as $\rho^s = \{\rho^s_1, \rho^s_2, . . . , \rho^s_n\}$. Given the dialogue context $c$, user persona $\rho^u$ and chatbot persona $\rho^s$, we are interested in finding the corresponding persona and predicting the chatbot response $s_t$.

To support our task, we collect and release a new dataset, entitled DuLeMon. In DuLeMon, the chatbot actively remembers and reasonably uses what the user has said about their persona while maintaining consistency in its persona, allowing the conversation to proceed more deeply. In a nutshell, our DuLeMon dataset has two essential features: During the conversation, the chatbot can see the persona of both parties; the other is that the persona associated with the response is explicitly annotated in our dataset. Unlike the PersonaChat dataset, the setting in DuLeMon is that one speaker plays the role of a chatbot, and the other plays the user's role. We elaborate on the construction process of the dataset as the following.

 \textbf{(1) Persona collection}: The persona is mainly from the translation and rewriting of persona in PersonaChat. The chatbot's persona is only visible to itself, and the chatbot can use its persona information to chat with the user, as shown in Figure \ref{fig:3}. The user's persona contains two parts: persona that the chatbot already knows and persona that the chatbot does not know. The first part is the user's persona that the chatbot has learned through historical conversations. This part is randomly selected from multiple personas of each user. The chatbot needs to use this information to guide the conversation during the chat process. It should be noted that in order to simulate the situation at the beginning of the chat, this part may be empty.
 
\textbf{(2) Dialogue collection}: For each dialogue, two crowd-workers (one plays the chatbot, the other plays the user) are randomly paired and given random persona. They are required to organize a dialogue based on the given persona. The chatbot should think more about chatting to make it go on. It should utilize the known user's persona to conduct the in-depth chat. The user will act as an ordinary user to cooperate with the conversation. The content of the chat can be selected from the given persona. It must not be irrelevant for the given information, nor can it conflict with the given persona.

\textbf{(3) Persona Grounding Labeling}: This part annotates whether the current response uses the given persona information and whether the current sentence is a persona sentence.  For each utterance, we first let the annotators label whether it uses persona or not. Furthermore, the annotator should label the grounding persona (from chatbot or user) being used in the response. Therefore, through this process, the direct relationship between the response and the persona can be given.  Then, for sentences that use the persona, we further annotate whether the utterance is a persona sentence or not.

To scale the amount of data, we also collected conversations where the user's persona was not visible to the bot, following the PersonaChat \citep{zhang-etal-2018-personalizing}. Finally, our DuLeMon dataset consists of two parts. In DeLeMon-SELF, the bot only knows its own persona,
while in DuLeMon-BOTH, it also knows part of the user's persona (as described above). The overall statistics of the DuLeMon are shown in Table \ref{duLeMon-statistics}. 

\begin{table}[]
\begin{tabular}{lll}
\toprule
\textbf{Category}             & \textbf{SELF} & \textbf{BOTH} \\
\midrule
\# Dialogues                  & 24500       & 3001        \\
\# Utterances                 & 400472      & 48522       \\
Avg. \# turns                 & 16.3        & 16.2        \\
Avg. length of utterances     & 19.7        & 21.2        \\
Avg. \# bot persona           & 4.0         & 4.0         \\
Avg. \# user persona (seen)   & 0           & 4.4         \\
Avg. \# user persona (unseen) & 4.0         & 1.3        \\
\bottomrule
\end{tabular}
\caption{\label{duLeMon-statistics}Statistics of DuLeMon.}
\end{table}





\section{Model Architecture}

\begin{figure*}[htp]
    \centering
    \includegraphics[width=16cm]{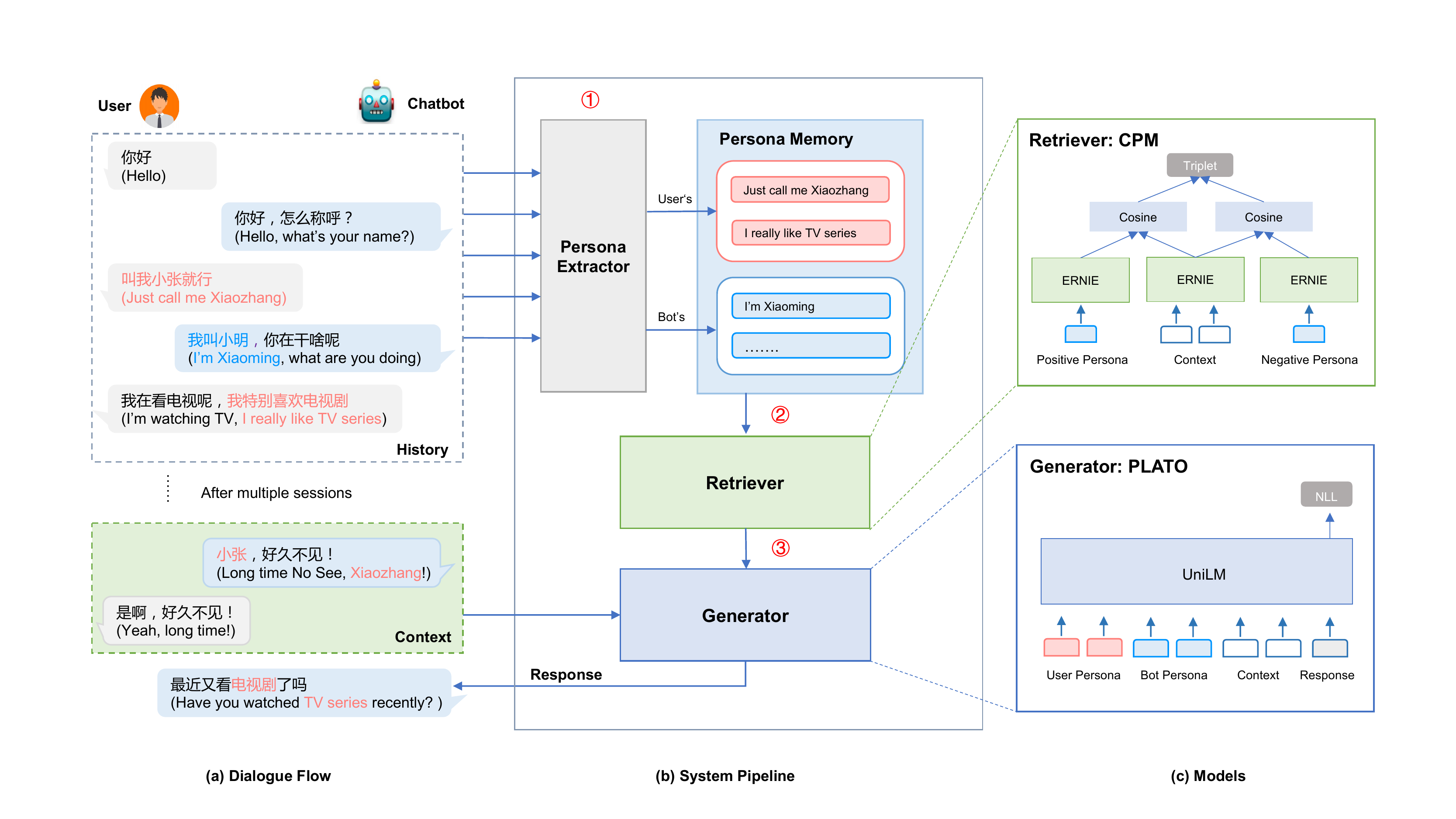}
    \caption{Illustration of our system PLATO-LTM. (a) shows the dialogue flow. (b) describes the modules and pipeline of our system. It consists of a persona extractor (PE), a long-term persona memory, a retriever, and a generator. \textcircled{1} The long-term memory contains both \textcolor{user_persona}{user persona} and \textcolor{bot_persona}{chatbot persona} extracted from the dialogue history by PE . \textcircled{2} The retriever uses context as query to retrieve related personas in memory \textcircled{3} concatenates the retrieved text to the context and use the generator to produce the generated response. (c) details our generator PLATO-2 and ranker CPM (Context Persona Matching).} 
    \label{fig:2}
\end{figure*}

In this work, we propose a long-term memory dialogue system based on an explicit memory read-write mechanism. It includes three parts: persona extractor, long-term persona memory, and generation module. Through the read and write operations of the long-term memory module, the user's and chatbot's persona can be stored, updated, and read. The overall framework is shown in Figure\ref{fig:2}. 


\subsection{Persona Extractor}

Given an utterance or text span as input, our persona extractor can assign each input  a label to indicate if it contains persona information. Here we train an ERNIE-CNN network architecture in a supervised way on an annotated persona-utterance dataset as this persona extractor. Specifically, the ERNIE-CNN network employs a pre-trained ERNIE\footnote{https://wenxin.baidu.com/} \cite{sun2019ernie} network for sentence representation, and another CNN model \citep{DBLP:conf/emnlp/Kim14} for classification.

\textbf{Training procedure.} First, we collect the first-version training dataset, in which there are 6k utterances (from the DuLeMon corpus and Chinese social forum corpus) being human-annotated with positive or negative class labels. Second, using the aforementioned dataset, we train five ERNIE-CNN network (with different pre-training parameter versions) based models (called pc-stage1). Third, we employ these five models to automatically annotate 1.4 million utterances with labels, where these utterances are collected from the DuLeMon and the online Chinese social forum. We then refine this augmented dataset as the final-version dataset with the following steps: (a) Given an utterance, if there are at least two of the above five models identifying it as a positive sample, then it is attached with a positive label, (b) otherwise it is attached with a negative label. Finally, we train the five models on the final-version dataset and select the one with the best performance as our persona extractor (named pc-stage2). 

\textbf{Inference procedure.} First, given an utterance, we segment it into clauses with the use of punctuation marks. Second, we use the persona extractor mentioned above to classify each clause with a label and then collect the clause with a positive label as persona sentences.


\subsection {Long-Term Memory}
The long-term memory (LTM) module maintains memories to store the historical persona information from the user and the chatbot, respectively. The most critical operations are reading and writing based on the context persona matching (CPM) model. We use context encoder $E_c(\cdot)$ to encode the current context $c$, and use persona encoder $E_\rho(\cdot)$ to encode the persona $\rho_i$.  $E(\cdot)$ is the encoder’s output on the first input token ($\texttt{[CLS]}$), corresponding to the input’s pooled representation.

The encoder $E_c$ and $E_\rho$ is initialized with the ERNIE model and then trained on our DuLeMon corpus. 
For each training sample, we define the positive persona as the persona used in the current user's utterance and the bot's response (including bot persona and user persona seen by bot), and the negative persona as the remaining persona of the current session.
Given context $c$, a positive persona $\rho^+$, and a negative persona $\rho^-$, we use triplet loss to tune the network as:
$$
\max \left(sim(c, \rho^+) - sim(c, \rho^-)+\alpha, 0\right)
$$

We set the margin $\alpha=0.2$ in our experiments. Below we describe the specific read and write process of the long-term memory module.

\textbf{Write}: We use the PE module to identify the persona in the dialogue history as the candidate information to be written. It needs to eliminate duplicates before writing. Specifically, calculate the cosine similarity with the persona in memory to get the most approximate persona $\rho_j$. When the similarity between $\rho_i$ and $\rho_j$ exceeds the given duplication threshold $s_{dup}$, replace $\rho_j$ in memory with $\rho_i$; otherwise, write $\rho_i$ directly into the memory. When writing to memory, save $\{\rho_i, E_\rho(\rho_i)\}$ pair for the subsequent reading. We measure the distance with the cosine similarity as:
\begin{equation}\label{sim_p_p}
sim(\rho_i, \rho_j) = \cos(E_\rho(\rho_i), E_\rho(\rho_j))
\end{equation}

\textbf{Read}: The reading process can be regarded as the retrieval process from memory. First, we use the efficient similarity search of dense vectors to select candidates. Then a matching model is utilized to score the relevance of the candidates to the current context.   The similarity between the context and the persona using cosine similarity:

\begin{equation}\label{sim_c_p}
sim(c, \rho_i) = \cos(E_c(c), E_\rho(\rho_i))
\end{equation}

The top $k$ persona candidates $\rho^u$ in the user memory and top $k$ candidates $\rho^s$ in the chatbot memory are used for response generation. To model persona sparsity in dialogue, we filter out the persona, whose similarity score is lower than the similarity threshold $s_c$.

\subsection{Generation Module}

We trained our model on the basis of the PLATO-2 \citep{DBLP:journals/corr/abs-2006-16779} architecture which adopts the generic transformer language model \citep{NIPS2017_3f5ee243} and leverages a stack of masked multi-head self-attention layers to train on massive dialogue data \footnote{There are two stages within the PLATO-2 model, the first stage conduct candidate responses generation and the second stage conduct responses selection. We only implement our work on the first stage of PLATO-2.}.

Given the conversation context $c =\{u_1, s_1, u_2, s_2, ..., u_{t-1}, s_{t-1}, u_t\}$, the corresponding user persona $\rho^u$ and chatbot persona $\rho^s$, the ground
truth response as $r=\{x_{m+1}, x_{m+2}, ..., x_N\}$, the conditional probability of $p(r|c, \rho^u, \rho^s)$ can be written as the product of a series of conditional probabilities:

\begin{equation}\label{eq1}
p(r|c, \rho^u, \rho^s) = \prod_{t}^{N}p(r_t|c, \rho^u, \rho^s, r_{<t})
\end{equation}

Therefore, we need to minimize the following
negative log-likelihood (NLL) loss:

\begin{equation}\label{eq2}
\begin{aligned}
\mathcal{L}_{NLL} &= -\mathbb{E} \log p(r|c, \rho^u, \rho^s) \\ 
 &= -\mathbb{E}\sum_{t=1}^{T}\log p(r_t|c, \rho^u, \rho^s, r_{<t})
\end{aligned}
\end{equation}
where $T$ is the length of the target response $r$ and
$r_{<t}$ denotes previously generated words. Since the
response generation is a uni-directional decoding
process, each token in the response only attends to
those before it. As for the context, bi-directional attention is enabled for better natural language understanding.

We added two strategies to distinguish different roles in the dialogue and prevent the confusing use of persona information.  

\begin{itemize}
        \item Role Embedding \cite{bao2021platoxl}: different role embedding is used to distinguish the persona of different chat parties, abbreviated role\_embed.
        \item Role Token: splicing "system persona" before the chatbot persona and "user persona" before the user persona, abbreviated role\_token. 
\end{itemize}

\section{Experiments}
In this section, we present the baselines, experiment settings, model comparisons, and results of experiments.

\subsection{Compared Methods}
As baselines, we select state-of-the-art methods to compare with our method.

\begin{itemize}
        \item PLATO-2 \citep{DBLP:journals/corr/abs-2006-16779}: The SOTA open-domain
dialogue model. 
        \item PLATO-FT: The PLATO-2 model fine-tuned on our proposed DuLeMon dataset. 
        \item PLATO-LTM: The PLATO-FT model with our proposed long-term memory (LTM).
        \item PLATO-LTM w/o PE: PLATO-LTM without the persona extractor (PE) module, which stores all  history utterances (user and bot separately) into memory without persona extraction.
    \end{itemize}
    
\subsection{Experiment Settings}

     
\noindent \textbf{Automatic Evaluation Metrics.} We use  Precision, Recall and F1 to evaluate the persona classification model. For the long-term memory module, we use the AUC and recall@k to evaluate the ranking model. We evaluate responses generated by the models using PPL, BLEU \citep{DBLP:conf/acl/PapineniRWZ02}, and F1 with reference to the human-annotated responses and DISTINCT-1/2 \citep{DBLP:conf/acl/ZhaoZE17}. 
More recently, \citet{DBLP:journals/corr/abs-2001-09977} has shown the correlation between perplexity and human judgment in open-domain chit-chat models.
    
\noindent \textbf{Human Evaluation Metrics.} In human evaluation, we employ three utterance-level metrics, including coherence, consistency, engagingness. Three crowd-sourcing workers are asked to score the response/dialogue quality on a scale of [0, 1, 2]. The higher score, the better. These criteria are discussed as follows:

\begin{itemize}
    \item Coherence: an utterance-level metric, measuring whether the response is relevant and consistent with the context.
    \item Consistency: an utterance-level metric, evaluating whether the response is consistent with the persona in the dialogue history.
    \item Engagingness: an utterance-level metric, assessing whether the annotator would like to talk with the speaker for each response in the long-term conversation.
\end{itemize}

\subsection{Results}
In this part, we first analyze the effects of each module and then analyze the results of the manual evaluation of our entire system, PLATO-LTM.

\subsubsection{Results of Persona Extractor}
 We measure the performance of the persona extractor. To measure the performance of different models, we manually annotated the test set (the number of test sets is 200). We select the best of the first and second stages. The result is shown in Table \ref{nlu-table}.  The pc-stage2 model is better than that of the pc-stage1 model. The F1 of the model exceeds 0.9, which shows that our model can effectively recognize the persona information from the dialogue history and ensure that the persona information can be correctly stored in the long-term memory. Therefore, the pc-stage2 model is adopted in our system to recognize the persona in the dialogue history.

\begin{table}
\centering
\small
\begin{tabular}{lllll}
\toprule
\textbf{Model} & \textbf{ACC} & \textbf{Precision} & \textbf{Recall} & \textbf{F1} \\
\midrule
 pc-stage1 & 0.91 & 0.96  & 0.84 & 0.90 \\
 pc-stage2 & 0.92 &  0.95  & 0.87 & 0.91 \\
\bottomrule
\end{tabular}
\caption{\label{nlu-table}
Comparison of two-stage models of our persona classifier.
}
\end{table}

\begin{table*}[]
\centering
\small
\begin{tabular}{lccccc}
\toprule
\textbf{Model}                    & \textbf{PPL} & \textbf{BLUE-1/2} & \textbf{DISTINT-1/2} & \textbf{F1} \\
\midrule

PLATO-FT  12L                     & 13.641 & 0.190/0.081       & 0.061/0.277          & 21.02 \\
PLATO-FT 12L + role\_embed          & 13.387 & 0.180/0.080       & 0.062/0.274          & 20.98 \\
PLATO-FT 12L + role\_token                   & 13.553 & 0.193/0.081       & 0.060/0.272          & 21.28 \\
PLATO-FT 12L + role\_embed + role\_token      & \textbf{13.377} & \textbf{0.194/0.081}       & 0.060/0.267          & \textbf{21.59} \\
\midrule
PLATO-FT 32L + role\_embed + role\_token  & 9.380  & 0.194/0.087       & 0.068/0.296          & 22.61 \\
\bottomrule
\end{tabular}
\caption{\label{nlg-eval}
Comparison of automatic evaluation metric results among different generative models.
}

\end{table*}

\begin{table*}
\small
\centering
\begin{tabular}{lccc}
\toprule[1pt]
\textbf{}      \textbf{Model}           & \textbf{Coherence}      &  \textbf{Consistency}  & \textbf{Engagingness} \\
\midrule[0.5pt]  
PLATO-2           & \textbf{1.70}  & 0.13 & 1.46    \\
PLATO-FT             &  1.59          & 0.40   & 1.40     \\
PLATO-LTM        & 1.67         & \textbf{0.87} & \textbf{1.54}  \\
PLATO-LTM  w/o PE & 1.57      & 0.49    & 1.43  \\ 
\bottomrule[1pt]
\end{tabular}
\caption{\label{human-eval}
Comparison of human evaluation metric results on self-chat dialogues among our model and baselines.
All the above generation models are 32L. The PLATO-FT is with role embedding and role token strategies.}
\end{table*}

\subsubsection{Selection of Generative Models}
The generative model utilizes the current context and persona information retrieved from long-term memory to generate the response. We first evaluate the effect of the CPM model on retrieval persona information. The AUC on the automatic test set is 0.76, recall@5 is 0.83, which shows that our model can efficiently retrieve relevant persona from the long-term memory.

The effect of the generative model reflects the model's ability to use the content of long-term memory to generate the response. Therefore, we select the best generative model to utilize better the retrieved persona information to generate. The result is shown in Table \ref{nlg-eval}. We use the 12L model to conduct experiments to compare different models. The experiment results show that PLATO-FT + role\_embed + role\_token is the best. Compared to PLATO-FT, the PPL can decrease to 13.377, showing that both strategies are effective. In order to further improve the model, we increased the model size and further trained with the 32L model. Experiment results have shown that the PPL of the 32L model is lower than the 12L model by 4.4 and F1 increased by 2.5, which can further improve the generative model. Therefore, PLATO-FT 32L + role\_embed + role\_token model is adopted in our system.

\subsubsection{Human Evaluation}
Self-chat has been widely used in the evaluation of dialogue systems (\citealp{li2016deep}; \citealp{DBLP:conf/eacl/RollerDGJWLXOSB21}; \citealp{DBLP:journals/corr/abs-2006-16779}), where the model plays the roles of both parties in the dialogue.
To better control variables, we use our proposed PLATO-LTM as a user simulator in our experiments and ask all chatbots (including PLATO-LTM) to chat separately with the user simulator. After that, the crowd-sourcing workers evaluate only the responses generated by the chatbots other than the simulator. The details are as follows.

Each chatbot chats with the user simulator for 10 episodes, each containing 4 long sessions, and each session contains 16 rounds.  As in \citet{DBLP:journals/corr/abs-2006-16779}, we do not impose any restrictions on the chats except for specifying session openings. We pre-select some session openings from the DuLeMon test set, start the interactive conversation with these openings, and ask the two bots to perform chats given the context.

The results are shown in Table \ref{human-eval}, from which we can get the following key results:

\textbf{(1) The long-Term Memory mechanism can significantly improve dialogue consistency.} As shown in Table \ref{human-eval}, in terms of dialogue consistency, our two models, PLATO-LTM and PLATO-FT, can achieve scores of 0.87 and 0.40, respectively, which is significantly better than the baseline model PLATO-2. Furthermore, when we compare the performance of PLATO-LTM with PLATO-FT, it can be seen that the use of Long-Term Memory and persona extractor can boost the performance of PLATO-FT with a relative improvement of 118\%. Moreover, the model of PLATO-LTM w/o PE can achieve a score of 0.49, which is still better than the PLATO-FT model. It indicates that long-term memory without a persona extractor is still effective in improving persona consistency. 
    
\textbf{(2) With the long-term memory mechanism, the use of persona extractor can significantly improve persona consistency and dialogue engagingness.} As shown in Table \ref{human-eval}, in terms of dialogue consistency,  the two models, PLATO-LTM (using PE) and PLATO-LTM w/o PE, can achieve scores of 0.87 and 0.49 respectively, indicating that the use of persona extractor can significantly improve dialogue consistency. In terms of dialogue engagingness, PLATO-LTM can obtain a score of 1.54, outperforming the baseline model PLATO-2. In addition, when we remove PE from PLATO-LTM, its performance drops from 1.54 (the score of PLATO-LTM) to 1.43 (that of PLATO-LTM w/o PE), indicating that the use of persona extractor can improve the performance of PLATO-FT.  

\textbf{(3)  Fine-tuning on the small-scale dataset will slightly hurt the performance of pre-trained dialogue models in dialogue coherence.} In terms of dialogue coherence, the PLATO-FT model (finetuned on our dataset) achieve a score of 1.59, which is lower than that of the baseline model PLATO (not finetuned on our dataset). The possible reason is that during the self-play procedure for system evaluation, their dialogs usually cover a wide range of topics, and then it is challenging to generate appropriate or coherent responses when given these open-domain topics in contexts. The finetuning procedure might hurt the capability of the pre-trained dialogue model in terms of response appropriateness or dialogue coherence, leading to the inferior performance of PLATO-LTM and its variants.




\section{Conclusion}
In this paper, We present a novel LeMon (Long-term Memory Conversation) task and then build the corresponding dataset DuLeMon, introducing long-term persona modelling into large-scale generative dialogue models. We further propose a Long-Term Memory (LTM) as a plug-in component of state-of-the-art large-scale generative dialogue models. LTM consists of user memory and chatbot memory, where the user memory is for understanding and memorizing persona information mentioned by the user, and the chatbot memory attempts to keep its persona information to be continuously updated over time. Experiment results show that our system PLATO-LTM can make effective use of both parties' persona information from dialogue history to enhance dialogue consistency and engagingness when conducting a long-term conversation. In the future, we will further study the possibility of using reinforcement learning with human feedback signals to help long-term conversation.
\section{Ethical Considerations}
We are sure that DuLeMon has been collected in a manner that is consistent with the terms of use of any sources and the intellectual property and privacy rights of the original authors of the texts. Meanwhile, our project is approved by an IRB. Finally, we also provide details on the characteristics of DuLeMon and steps taken to ensure the potential problems with the quality of the dataset do not create additional risks.

\bibliography{acl}

\begin{thebibliography}{27}
\expandafter\ifx\csname natexlab\endcsname\relax\def\natexlab#1{#1}\fi

\bibitem[{Adiwardana et~al.(2020)Adiwardana, Luong, So, Hall, Fiedel,
  Thoppilan, Yang, Kulshreshtha, Nemade, Lu, and
  Le}]{DBLP:journals/corr/abs-2001-09977}
Daniel Adiwardana, Minh{-}Thang Luong, David~R. So, Jamie Hall, Noah Fiedel,
  Romal Thoppilan, Zi~Yang, Apoorv Kulshreshtha, Gaurav Nemade, Yifeng Lu, and
  Quoc~V. Le. 2020.
\newblock \href {http://arxiv.org/abs/2001.09977} {Towards a human-like
  open-domain chatbot}.
\newblock \emph{CoRR}, abs/2001.09977.

\bibitem[{Bang et~al.(2015)Bang, Noh, Kim, and Lee}]{7072837}
Jeesoo Bang, Hyungjong Noh, Yonghee Kim, and Gary~Geunbae Lee. 2015.
\newblock \href {https://doi.org/10.1109/35021BIGCOMP.2015.7072837}
  {Example-based chat-oriented dialogue system with personalized long-term
  memory}.
\newblock In \emph{2015 International Conference on Big Data and Smart
  Computing (BIGCOMP)}, pages 238--243.

\bibitem[{Bao et~al.(2020)Bao, He, Wang, Wu, Wang, Wu, Guo, Liu, and
  Xu}]{DBLP:journals/corr/abs-2006-16779}
Siqi Bao, Huang He, Fan Wang, Hua Wu, Haifeng Wang, Wenquan Wu, Zhen Guo,
  Zhibin Liu, and Xinchao Xu. 2020.
\newblock \href {http://arxiv.org/abs/2006.16779} {{PLATO-2:} towards building
  an open-domain chatbot via curriculum learning}.
\newblock \emph{CoRR}, abs/2006.16779.

\bibitem[{Bao et~al.(2021)Bao, He, Wang, Wu, Wang, Wu, Wu, Guo, Lu, Huang,
  Tian, Xu, Lin, and Niu}]{bao2021platoxl}
Siqi Bao, Huang He, Fan Wang, Hua Wu, Haifeng Wang, Wenquan Wu, Zhihua Wu, Zhen
  Guo, Hua Lu, Xinxian Huang, Xin Tian, Xinchao Xu, Yingzhan Lin, and Zhengyu
  Niu. 2021.
\newblock \href {http://arxiv.org/abs/2109.09519} {Plato-xl: Exploring the
  large-scale pre-training of dialogue generation}.

\bibitem[{Campos et~al.(2018)Campos, Kennedy, and
  Lehman}]{10.5555/3237383.3237945}
Joana Campos, James Kennedy, and Jill~F. Lehman. 2018.
\newblock Challenges in exploiting conversational memory in human-agent
  interaction.
\newblock In \emph{Proceedings of the 17th International Conference on
  Autonomous Agents and MultiAgent Systems}, AAMAS '18, page 1649–1657,
  Richland, SC. International Foundation for Autonomous Agents and Multiagent
  Systems.

\bibitem[{Dinan et~al.(2019)Dinan, Logacheva, Malykh, Miller, Shuster, Urbanek,
  Kiela, Szlam, Serban, Lowe, Prabhumoye, Black, Rudnicky, Williams, Pineau,
  Burtsev, and Weston}]{DBLP:journals/corr/abs-1902-00098}
Emily Dinan, Varvara Logacheva, Valentin Malykh, Alexander~H. Miller, Kurt
  Shuster, Jack Urbanek, Douwe Kiela, Arthur Szlam, Iulian Serban, Ryan Lowe,
  Shrimai Prabhumoye, Alan~W. Black, Alexander~I. Rudnicky, Jason Williams,
  Joelle Pineau, Mikhail~S. Burtsev, and Jason Weston. 2019.
\newblock \href {http://arxiv.org/abs/1902.00098} {The second conversational
  intelligence challenge (convai2)}.
\newblock \emph{CoRR}, abs/1902.00098.

\bibitem[{Elvir et~al.(2017)Elvir, Gonzalez, Walls, and
  Wilder}]{ElvirGonzalezWallsWilder+2017+1+21}
Miguel Elvir, Avelino~J. Gonzalez, Christopher Walls, and Bryan Wilder. 2017.
\newblock \href {https://doi.org/doi:10.1515/jisys-2015-0094} {Remembering a
  conversation – a conversational memory architecture for embodied
  conversational agents}.
\newblock \emph{Journal of Intelligent Systems}, 26(1):1--21.

\bibitem[{Huang et~al.(2020)Huang, Zhu, and Gao}]{10.1145/3383123}
Minlie Huang, Xiaoyan Zhu, and Jianfeng Gao. 2020.
\newblock \href {https://doi.org/10.1145/3383123} {Challenges in building
  intelligent open-domain dialog systems}.
\newblock \emph{ACM Trans. Inf. Syst.}, 38(3).

\bibitem[{Kim et~al.(2014)Kim, Bang, Choi, Ryu, Koo, and
  Lee}]{DBLP:conf/interspeech/KimBCRKL14}
Yonghee Kim, Jeesoo Bang, Junhwi Choi, Seonghan Ryu, Sangjun Koo, and
  Gary~Geunbae Lee. 2014.
\newblock \href {https://doi.org/10.1007/978-3-319-15557-9\_8} {Acquisition and
  use of long-term memory for personalized dialog systems}.
\newblock In \emph{Multimodal Analyses enabling Artificial Agents in
  Human-Machine Interaction - Second International Workshop, {MA3HMI} 2014,
  Held in Conjunction with {INTERSPEECH} 2014, Singapore, Singapore, September
  14, 2014, Revised Selected Papers}, volume 8757 of \emph{Lecture Notes in
  Computer Science}, pages 78--87. Springer.

\bibitem[{Kim(2014)}]{DBLP:conf/emnlp/Kim14}
Yoon Kim. 2014.
\newblock \href {https://doi.org/10.3115/v1/d14-1181} {Convolutional neural
  networks for sentence classification}.
\newblock In \emph{Proceedings of the 2014 Conference on Empirical Methods in
  Natural Language Processing, {EMNLP} 2014, October 25-29, 2014, Doha, Qatar,
  {A} meeting of SIGDAT, a Special Interest Group of the {ACL}}, pages
  1746--1751. {ACL}.

\bibitem[{Kingma and Ba(2015)}]{DBLP:journals/corr/KingmaB14}
Diederik~P. Kingma and Jimmy Ba. 2015.
\newblock \href {http://arxiv.org/abs/1412.6980} {Adam: {A} method for
  stochastic optimization}.
\newblock In \emph{3rd International Conference on Learning Representations,
  {ICLR} 2015, San Diego, CA, USA, May 7-9, 2015, Conference Track
  Proceedings}.

\bibitem[{Li et~al.(2016{\natexlab{a}})Li, Galley, Brockett, Spithourakis, Gao,
  and Dolan}]{DBLP:conf/acl/LiGBSGD16}
Jiwei Li, Michel Galley, Chris Brockett, Georgios~P. Spithourakis, Jianfeng
  Gao, and William~B. Dolan. 2016{\natexlab{a}}.
\newblock \href {https://doi.org/10.18653/v1/p16-1094} {A persona-based neural
  conversation model}.
\newblock In \emph{Proceedings of the 54th Annual Meeting of the Association
  for Computational Linguistics, {ACL} 2016, August 7-12, 2016, Berlin,
  Germany, Volume 1: Long Papers}. The Association for Computer Linguistics.

\bibitem[{Li et~al.(2016{\natexlab{b}})Li, Monroe, Ritter, Galley, Gao, and
  Jurafsky}]{li2016deep}
Jiwei Li, Will Monroe, Alan Ritter, Michel Galley, Jianfeng Gao, and Dan
  Jurafsky. 2016{\natexlab{b}}.
\newblock \href {http://arxiv.org/abs/1606.01541} {Deep reinforcement learning
  for dialogue generation}.

\bibitem[{Lim(2012)}]{Lim2012}
Mei~Yii Lim. 2012.
\newblock \href {https://doi.org/10.1007/978-3-642-25691-2_10} {\emph{Memory
  Models for Intelligent Social Companions}}, pages 241--262. Springer Berlin
  Heidelberg, Berlin, Heidelberg.

\bibitem[{Lin et~al.(2020)Lin, Liu, Winata, Cahyawijaya, Madotto, Bang, Ishii,
  and Fung}]{DBLP:journals/corr/abs-2003-07568}
Zhaojiang Lin, Zihan Liu, Genta~Indra Winata, Samuel Cahyawijaya, Andrea
  Madotto, Yejin Bang, Etsuko Ishii, and Pascale Fung. 2020.
\newblock \href {http://arxiv.org/abs/2003.07568} {Xpersona: Evaluating
  multilingual personalized chatbot}.
\newblock \emph{CoRR}, abs/2003.07568.

\bibitem[{Mazar{\'e} et~al.(2018)Mazar{\'e}, Humeau, Raison, and
  Bordes}]{mazare-etal-2018-training}
Pierre-Emmanuel Mazar{\'e}, Samuel Humeau, Martin Raison, and Antoine Bordes.
  2018.
\newblock \href {https://doi.org/10.18653/v1/D18-1298} {Training millions of
  personalized dialogue agents}.
\newblock In \emph{Proceedings of the 2018 Conference on Empirical Methods in
  Natural Language Processing}, pages 2775--2779, Brussels, Belgium.
  Association for Computational Linguistics.

\bibitem[{Papineni et~al.(2002)Papineni, Roukos, Ward, and
  Zhu}]{DBLP:conf/acl/PapineniRWZ02}
Kishore Papineni, Salim Roukos, Todd Ward, and Wei{-}Jing Zhu. 2002.
\newblock \href {https://doi.org/10.3115/1073083.1073135} {Bleu: a method for
  automatic evaluation of machine translation}.
\newblock In \emph{Proceedings of the 40th Annual Meeting of the Association
  for Computational Linguistics, July 6-12, 2002, Philadelphia, PA, {USA}},
  pages 311--318. {ACL}.

\bibitem[{Qian et~al.(2018)Qian, Huang, Zhao, Xu, and Zhu}]{ijcai2018-595}
Qiao Qian, Minlie Huang, Haizhou Zhao, Jingfang Xu, and Xiaoyan Zhu. 2018.
\newblock \href {https://doi.org/10.24963/ijcai.2018/595} {Assigning
  personality/profile to a chatting machine for coherent conversation
  generation}.
\newblock In \emph{Proceedings of the Twenty-Seventh International Joint
  Conference on Artificial Intelligence, {IJCAI-18}}, pages 4279--4285.
  International Joint Conferences on Artificial Intelligence Organization.

\bibitem[{Roller et~al.(2021)Roller, Dinan, Goyal, Ju, Williamson, Liu, Xu,
  Ott, Smith, Boureau, and Weston}]{DBLP:conf/eacl/RollerDGJWLXOSB21}
Stephen Roller, Emily Dinan, Naman Goyal, Da~Ju, Mary Williamson, Yinhan Liu,
  Jing Xu, Myle Ott, Eric~Michael Smith, Y{-}Lan Boureau, and Jason Weston.
  2021.
\newblock \href {https://www.aclweb.org/anthology/2021.eacl-main.24/} {Recipes
  for building an open-domain chatbot}.
\newblock In \emph{Proceedings of the 16th Conference of the European Chapter
  of the Association for Computational Linguistics: Main Volume, {EACL} 2021,
  Online, April 19 - 23, 2021}, pages 300--325. Association for Computational
  Linguistics.

\bibitem[{Sun et~al.(2019)Sun, Wang, Li, Feng, Tian, Wu, and
  Wang}]{sun2019ernie}
Yu~Sun, Shuohuan Wang, Yukun Li, Shikun Feng, Hao Tian, Hua Wu, and Haifeng
  Wang. 2019.
\newblock \href {http://arxiv.org/abs/1907.12412} {Ernie 2.0: A continual
  pre-training framework for language understanding}.

\bibitem[{Vaswani et~al.(2017)Vaswani, Shazeer, Parmar, Uszkoreit, Jones,
  Gomez, Kaiser, and Polosukhin}]{NIPS2017_3f5ee243}
Ashish Vaswani, Noam Shazeer, Niki Parmar, Jakob Uszkoreit, Llion Jones,
  Aidan~N Gomez, \L~ukasz Kaiser, and Illia Polosukhin. 2017.
\newblock \href
  {https://proceedings.neurips.cc/paper/2017/file/3f5ee243547dee91fbd053c1c4a845aa-Paper.pdf}
  {Attention is all you need}.
\newblock In \emph{Advances in Neural Information Processing Systems},
  volume~30. Curran Associates, Inc.

\bibitem[{Xu et~al.(2021)Xu, Szlam, and Weston}]{xu2021goldfish}
Jing Xu, Arthur Szlam, and Jason Weston. 2021.
\newblock \href {http://arxiv.org/abs/2107.07567} {Beyond goldfish memory:
  Long-term open-domain conversation}.

\bibitem[{Zhang et~al.(2018)Zhang, Dinan, Urbanek, Szlam, Kiela, and
  Weston}]{zhang-etal-2018-personalizing}
Saizheng Zhang, Emily Dinan, Jack Urbanek, Arthur Szlam, Douwe Kiela, and Jason
  Weston. 2018.
\newblock \href {https://doi.org/10.18653/v1/P18-1205} {Personalizing dialogue
  agents: {I} have a dog, do you have pets too?}
\newblock In \emph{Proceedings of the 56th Annual Meeting of the Association
  for Computational Linguistics (Volume 1: Long Papers)}, pages 2204--2213,
  Melbourne, Australia. Association for Computational Linguistics.

\bibitem[{Zhang et~al.(2017)Zhang, Liu, Wang, and
  Zhu}]{DBLP:journals/corr/ZhangLWZ17}
Weinan Zhang, Ting Liu, Yifa Wang, and Qingfu Zhu. 2017.
\newblock \href {http://arxiv.org/abs/1701.02073} {Neural personalized response
  generation as domain adaptation}.
\newblock \emph{CoRR}, abs/1701.02073.

\bibitem[{Zhao et~al.(2017)Zhao, Zhao, and
  Esk{\'{e}}nazi}]{DBLP:conf/acl/ZhaoZE17}
Tiancheng Zhao, Ran Zhao, and Maxine Esk{\'{e}}nazi. 2017.
\newblock \href {https://doi.org/10.18653/v1/P17-1061} {Learning
  discourse-level diversity for neural dialog models using conditional
  variational autoencoders}.
\newblock In \emph{Proceedings of the 55th Annual Meeting of the Association
  for Computational Linguistics, {ACL} 2017, Vancouver, Canada, July 30 -
  August 4, Volume 1: Long Papers}, pages 654--664. Association for
  Computational Linguistics.

\bibitem[{Zheng et~al.(2019)Zheng, Chen, Huang, Liu, and
  Zhu}]{DBLP:journals/corr/abs-1901-09672}
Yinhe Zheng, Guanyi Chen, Minlie Huang, Song Liu, and Xuan Zhu. 2019.
\newblock \href {http://arxiv.org/abs/1901.09672} {Personalized dialogue
  generation with diversified traits}.
\newblock \emph{CoRR}, abs/1901.09672.

\bibitem[{Zhong et~al.(2020)Zhong, Zhang, Wang, Liu, and
  Miao}]{zhong-etal-2020-towards}
Peixiang Zhong, Chen Zhang, Hao Wang, Yong Liu, and Chunyan Miao. 2020.
\newblock \href {https://doi.org/10.18653/v1/2020.emnlp-main.531} {Towards
  persona-based empathetic conversational models}.
\newblock In \emph{Proceedings of the 2020 Conference on Empirical Methods in
  Natural Language Processing (EMNLP)}, pages 6556--6566, Online. Association
  for Computational Linguistics.

\end{thebibliography}
\bibliographystyle{acl_natbib}

\appendix

\section{Details of Data Collection}
\label{sec:appendix1}
The collection processes of DuLeMon are as follows.

\begin{itemize}
       \item The crowdworkers enter the chat interface in pairs, and role 1 initiates a conversation;
       \item The chat content can include opening greetings, self-introduction, chatting content that conforms to the persona information, asking the other party's questions, answering the other's questions, and so on. The information used in the chat must be consistent with the given personal information;
       \item The dialogue contains at least 8 turns (each person speaks at least 8 utterances);
   \end{itemize}

At the same time, we also let the crowdworkers pay attention to the follows: 1. Use as many words as possible, and do not repeat them. The overall dialogue strives to be natural, smooth, and not embarrassing. 2. Do not simply copy and paste the sentences in the personal information and express them as richly as possible. If it is found that 50\% of the fragments of any given sentence appear in the conversation, it is a non-compliant conversation. 3. When using persona information, do not copy it entirely, and talk about relevant content around the persona. For example, if the persona setting contains the sentence \textit{"I am a painter"}, the response can be that \textit{"I have painted many beautiful paintings and held several exhibitions"}; 4. If the question raised by the other speaker is not covered in the given personal information, the reply can be freely used; if there is any reference or related information in the given personal information, reply according it.

\section{Details of Models}
\label{sec:appendix2}
\textbf{Generation Model} For the Generation model, We follow PLATO-2 \citep{DBLP:journals/corr/abs-2006-16779}. The maximum length of context, user persona, and chatbot persona are set to 384, 76, and 52, respectively. The vocabulary contains 30K Chinese BPE tokens. We optimize all models using Adam \citep{DBLP:journals/corr/KingmaB14} with every batch of $B=16384$ tokens and learning rate of $lr=5e-5$. We conduct all experiments on NVIDIA V100 32GB and A100 48GB GPUs.\\
\textbf{Long-term Memory} For both user memory and chatbot memory, we set duplication threshold $s_{dup}=0.95$, number of candidates $K=5$, and similarity threshold $s_c=0.7$. Due to the persona sparsity of dialogue and the efficiency of our persona storage, we do not limit the memory capacity.

\section{Cases of PLATO-LTM}
\label{sec:appendix3}
\begin{figure*}[htp]
    \centering
    \includegraphics[width=16cm]{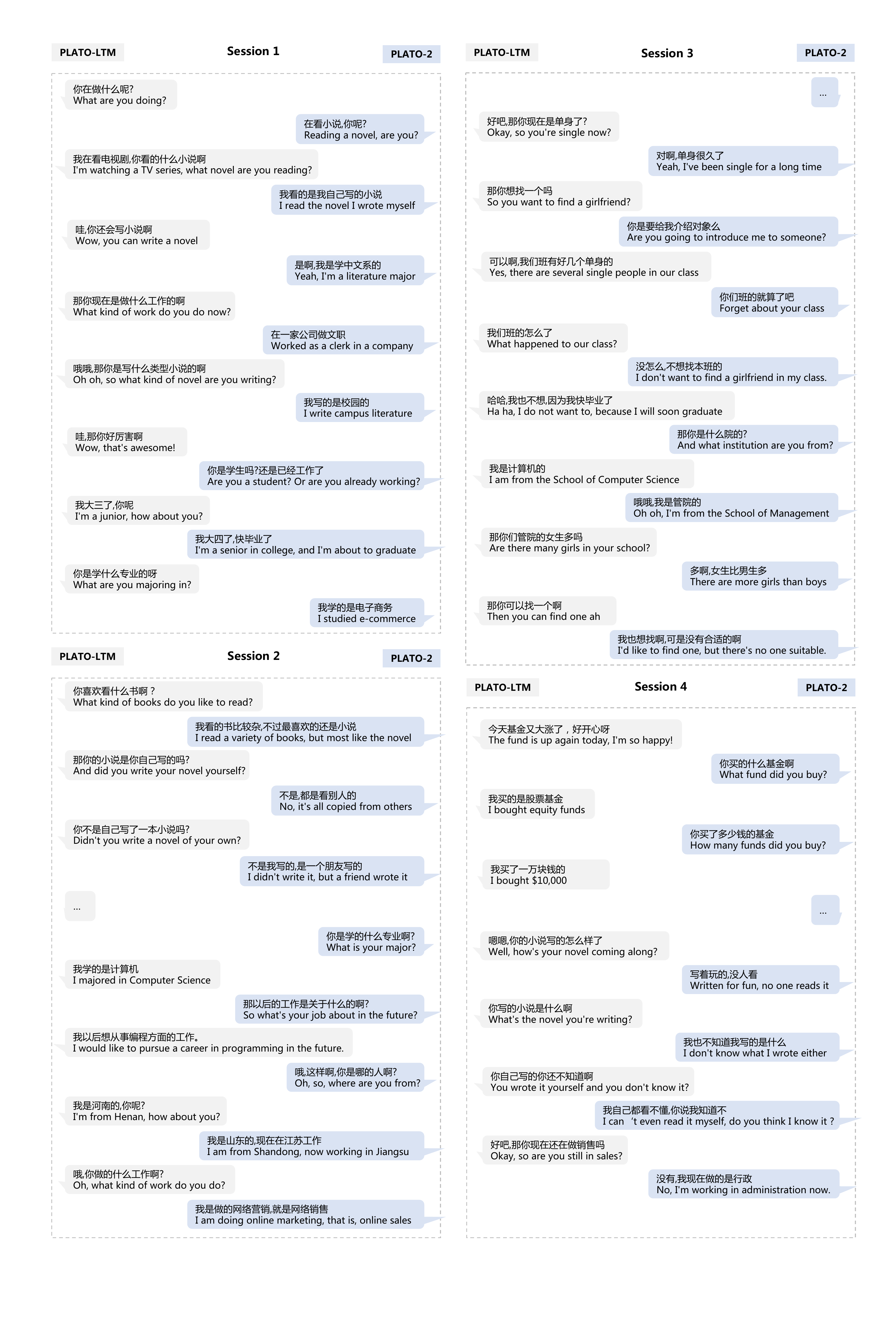}
    \caption{A cherry-picked example of one episode conversation between PLATO-LTM and PLATO-2.} 
    \label{fig:4}
\end{figure*}

To concretely demonstrate the long-term persona ability in a long-term conversation, we further provide a cherry-picked example of one episode conversation (between PLATO-LTM and PLATO-2) in Figure \ref{fig:4}. 

\end{document}